\documentclass[runningheads]{llncs}
\usepackage{graphicx}
\usepackage{hyperref}
\usepackage{multirow} 
\usepackage{url}
\usepackage{wrapfig}
\usepackage[frozencache=true,cachedir=minted-cache]{minted} 

\begin{document}
\title{Scoping Review of Active Learning Strategies and their Evaluation Environments for Entity Recognition Tasks}
\titlerunning{Scoping Review: Active Learning for Entity Recognition}
%
 \author{Philipp Kohl\inst{1}\orcidID{0000-0002-5972-8413} \and Yoka Krämer\inst{1}\orcidID{0009-0006-7326-3268} \and Claudia Fohry\inst{2}
 \and Bodo Kraft\inst{1}
 }
 \authorrunning{P. Kohl et al.}
\institute{FH Aachen -- University of Applied Sciences, 52428 Jülich, Germany \email{\{p.kohl,y.kraemer,kraft\}@fh-aachen.de \and
University of Kassel, 34121 Kassel, Germany
\email{fohry@uni-kassel.de}}\\}
%
\maketitle              
\begin{abstract}

We conducted a scoping review for active learning in the domain of natural language processing (NLP), which we summarize in accordance with the PRISMA-ScR guidelines as follows:

\textbf{Objective:} Identify active learning strategies that were proposed for entity recognition and their evaluation environments (datasets, metrics, hardware, execution time).

\textbf{Design:} We used Scopus and ACM as our search engines. We compared the results with two literature surveys to assess the search quality. We included peer-reviewed English publications introducing or comparing active learning strategies for entity recognition. 

\textbf{Results:} We analyzed 62 relevant papers and identified 106 active learning strategies. We grouped them into three categories: exploitation-based (60x), exploration-based (14x), and hybrid strategies (32x). We found that all studies used the F1-score as an evaluation metric. Information about hardware (6x) and execution time (13x) was only occasionally included. The 62 papers used 57 different datasets to evaluate their respective strategies. Most datasets contained newspaper articles or biomedical/medical data. Our analysis revealed that 26 out of 57 datasets are publicly accessible.

\textbf{Conclusion:} 
Numerous active learning strategies have been identified, along with significant open questions that still need to be addressed. Researchers and practitioners face difficulties when making data-driven decisions about which active learning strategy to adopt. Conducting comprehensive empirical comparisons using the evaluation environment proposed in this study could help establish best practices in the domain.

\end{abstract}

\section{Introduction}
\label{sec:intro}
Recent years showed significant advancements \cite{vaswaniAttentionAllYou2023,brownLanguageModelsAre2020,devlinBERTPretrainingDeep2018} in natural language processing (NLP): Large language models (LLMs) emerged \cite{brown_language_2020}, facilitating new methodologies by describing tasks in natural language without a strong formalism. Because resource-intensive LLMs are not always superior \cite{zhuo_red_2023}, smaller, supervised learning-based models are still highly relevant for specialized domains or use cases that require rapid inference or are constrained by hardware limitations (such as mobile devices or offline scenarios) \cite{jayakumar_large_2023}.

One of these domains is entity recognition \cite{settles_analysis_2008}. \textit{Entity recognition} ~(ER) describes the task of assigning a label to a sequence of words (e.g. to extract a person, a date or any other predefined label). To apply supervised learning to ER, data must be annotated. The manual annotation process, in which humans annotate data points with these predefined labels, is time-intensive and expensive \cite{zhang_survey_2022}. Its output is an annotated dataset, which is also called \textit{corpus} (pl. \textit{corpora}) in the NLP domain. We use the terms interchangeably in this paper. 
 
Researchers have been exploring supporting methods to reduce the annotation effort, such as semi-supervised learning \cite{tran_hybrid_2017,hassanzadeh_two-phase_2013}, self-learning \cite{tran_combination_2017,zhong_chinese_2014}, and active learning \cite{arora_active_2007,settles_active_2009}. \textit{Active learning} (AL) is an approach to strategically or heuristically select data points for human annotation. This methodology can reduce the number of data points required to achieve competitive model performance compared to the classical annotation and training process. Thus, AL can decrease the time and cost of the annotation and training processes. However, the selection of an appropriate AL strategy is crucial. Selecting an inappropriate strategy can lead to lower performance compared to random data selection \cite{chang_using_2020,kohl_ale_2023}.

Over the past two decades, researchers have developed many active learning strategies in the field of NLP for various scenarios. However, it is still challenging for researchers and practitioners to select a promising strategy for a given use case. While existing AL surveys provide taxonomies \cite{settles_active_2009,zhan_comparative_2022,zhang_survey_2022}, there is still a lack of comprehensive performance analyses. Towards closing this gap, an overview of the domain can support researchers in conducting such analyses.
Therefore, we executed a scoping review focusing on active learning strategies and their evaluation environments limited to the entity recognition task in NLP. We concentrated our review on model-agnostic strategies so researchers can use our results for a broad range of models. Our review answers the following \textit{review questions}:
\begin{enumerate} 
    \item \textit{Which model-agnostic AL strategies have been applied to ER?}
    \item \textit{How did the researchers evaluate their strategies?}
    \begin{enumerate}
        \item \textit{Which datasets did they use?}
        \item \textit{Which metrics did they use to compare AL strategies?}
        \item \textit{How much time do the AL strategies need for initialization, proposing new data points to annotators, and model retraining (in case of exploitation) depending on the hardware?}
    \end{enumerate}
\end{enumerate}

We chose the ER task due to its complexity in the annotation process \cite{culotta_corrective_2006} and AL \cite{settles_active_2009}. The complexity results from the ER model, which makes decisions for every token (e.g., word). Many AL strategies ($>$ 80) compute the relevance of a data point based on these individual decisions. 

For our review, we selected the format of a scoping review \cite{grant_typology_2009,munnSystematicReviewScoping2018} because we give an overview of the domain by identifying the research field's available strategies, datasets, metrics, execution times, and hardware used. We also identified research gaps and open questions, which enables other researchers to conduct a systematic review with precisely defined research questions in the field of AL and ER based on our work\footnote{The preparation of systematic reviews is a strong hint for performing a scoping review.}.

We adhere closely to the PRISMA-ScR \cite{triccoPRISMAExtensionScoping2018} reporting schema and checklist. The schema provides detailed guidelines for conducting a scoping review (databases, criteria, search, data charting, ...) and writing a paper with all necessary information.
We publish our additional materials (exhaustive lists of all information regarding the review questions) publicly on GitHub\footnote{\url{https://github.com/philipp-kohl/scoping-review-active-learning-er}}.

\autoref{sec:definitions} defines the entity recognition task and active learning and outlines established taxonomies. \autoref{sec:scoping-review} details our review process, ensuring reproducibility and extensibility. \autoref{sec:results} presents the aggregated findings on AL strategies and their evaluation environments alongside open research questions. \autoref{sec:related-work} discusses related work on reducing annotation efforts besides AL.
\autoref{sec:ethical} addresses ethical considerations, and we conclude by summarizing our findings and highlighting future directions in \autoref{sec:conclusion}.

\section{Definitions}
\label{sec:definitions}
Our scoping review focuses on the application of AL to ER. These terms are not used uniformly in the literature.
To make our work reproducible and comprehensible, this section starts with precise definitions of both concepts concentrated on the NLP domain, as they are assumed throughout this paper. We also use the definitions as eligibility criteria (\autoref{sec:eligibility}) for our scoping review.

\subsection{Entity Recognition}
\label{sec:def-er}
Entity recognition (ER) describes the NLP task of using a machine learning model to find entities automatically (e.g., persons or organizations) in a text. ER works with an arbitrary predefined label set. A specialized type of ER is \textit{named entity recognition} \cite{sharma_named_2022}, which focuses on proper nouns with a label set of person, organization, location, and dates. 

ER splits the text into tokens (in a simplified format into words) and assigns a label to each token. This type is called sequence labeling. The literature distinguishes between sequence \cite{sharma_named_2022} and span labeling approaches \cite{son_jointly_2022,zaratiana_gnner_2022}. For sequence labeling approaches, it is challenging to label overlapping entities because every token receives only one label. Span labeling closes this gap. Spans represent $n$ consecutive tokens. Span labeling enumerates all spans with length $1 - n$ and classifies each span. Thus, span labeling has to make more decisions, which makes it more complex.

In this paper, we use the term \textit{entity recognition} for sequence and span labeling approaches with an arbitrary predefined label set.

\subsection{Active Learning}
\label{sec:def-al}

\begin{figure}[htb]
    \centering
    \includegraphics[width=0.7\linewidth]{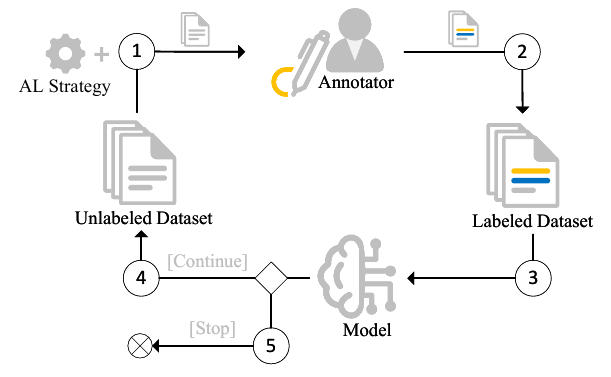}
    \caption{Flow diagram for the pool-based active learning cycle following \cite{settles_active_2009,kohl_ale_2023}.}
    \label{fig:al-cycle}
\end{figure}

Active learning (AL) reduces the annotation effort by selecting data points from an unannotated corpus with an AL strategy. \autoref{fig:al-cycle} shows the \textit{AL cycle}: 1) We start with a pool of unlabeled data points. The AL strategy selects data points from the pool and passes them to the human annotator. 2) Once the annotator has enriched the data points with the labels, the new labeled batch is added to the already labeled dataset. 3) When the amount of newly added data points reaches a threshold, the (re-)training of the NLP model will be triggered. 4) A new annotation cycle is started if no stopping criterion\footnote{Stopping criteria are e.g., desired model performance level or the unavailability of unlabeled data points.} is fulfilled. 5) This cycle repeats until a stopping criterion is met.

AL assumes that data points are not equally valuable for the amount of knowledge the NLP model gains \cite{ren_survey_2022}. AL algorithms pick the data points for annotating and, therefore, also for training to maximize the knowledge gained per annotated data point \cite{ren_survey_2022,mironczuk_recent_2018,zhou_progress_2020}. Ideally, this process reduces annotation effort and cost compared to sequential or random data selection \cite{zhou_progress_2020,ren_survey_2022}. However, using an unsuitable strategy can even lower performance compared to random selection \cite{chang_using_2020,kohl_ale_2023}.

We address pool-based AL, where we have a (large) unlabeled dataset, and select the most promising data points by the strategy. An alternative approach is stream-based AL \cite{loy_stream-based_2012}. The data is passed one by one to the strategy. The strategy decides whether to propose the single data point to the annotator without incorporating information from other data points. Users apply stream-based AL e.g., due to limited hardware settings.

As a basis for our own work, we adopted the well-known taxonomies from surveys \cite{mendonca_query_2020,bondu_exploration_2010,zhang_survey_2022,settles_active_2009} to categorize active learning strategies. Thereby, we modified the terminology to consistently use the same term for semantically similar concepts, improving our work's clarity and readability. 
On the top level of our categorization of AL strategies, we follow the distinction into \textit{exploitation-based}, \textit{exploration-based}, and \textit{hybrid} strategies:

\textbf{Exploitation methods} leverage feedback from the ER model to assess the potential value of a data point in the learning process \cite{mendonca_query_2020}. These methods typically use uncertainty scores, disagreement among multiple weak learners, or performance predictions as their basis. Exploitation strategies' intuition wants to enhance the decision boundaries, although they tend to focus on outliers \cite{zhang_survey_2022}. 
    The authors of \cite{zhang_survey_2022} use the term \textit{informativeness}, which is very similar to the understanding of exploitation. Hence, we combine the terms to have a consistent naming throughout this work. Informativeness strategies consider a single independent instance without assessing the relation to other data points. They do not explicitly state the need for model feedback, although all the stated strategies use model information. Thus, the definition aligns well with \textit{exploitation}.

In the case of AL for ER, we have to consider that ER works on token- or span-level. This gives us single feedback information for each token, which AL strategies have to interpret to measure the usefulness of the whole data point. For this purpose, different \textit{aggregation methods} are known in the literature \cite{sapci_focusing_2023}: e.g., total sum, average, or single most uncertainty.

\textbf{Exploration methods} select data points independently of model feedback. They use vector representations combined with clustering approaches based on density, diversity, and discriminative to determine a data point's relevance. Exploration strategies aim to cover the vector space 
holistically \cite{mendonca_query_2020}.
    \cite{zhang_survey_2022} uses the term \textit{representativeness} for exploration. Similar to exploitation and informativeness, we use the term exploration in this work. The authors use representativeness for strategies that include information on multiple data points to select a subset.

\textbf{Hybrid methods} combine exploitation and exploration methods \cite{zhang_survey_2022}. They use several strategies sequentially, in parallel, or combine them in weighted aggregation. This way, strategies compensate for other strategies' drawbacks (e.g., selecting outliers). Because hybrid strategies incorporate exploitation strategies, we must also consider aggregation methods.

We use this taxonomy to group the AL strategies of our screened papers in \autoref{sec:al-strats}.

\section{Methodology}
\label{sec:scoping-review}
Our scoping review follows the procedure proposed by PRISMA-ScR \cite{triccoPRISMAExtensionScoping2018}. We have selected and analyzed papers introducing or modifying active learning strategies applied to ER. For these papers, we list and group the active learning strategies and evaluation datasets. We inspected the evaluation environment to see which metrics researchers use and if they describe the used hardware and give information about execution times. 

\subsection{Search Engines} \label{sec:search}

We based the search engine selection on \cite{gusenbauer_which_2020}.
We chose \textit{Scopus}\footnote{\url{https://www.scopus.com/}} as our primary search engine because of its literature coverage, advanced searching, and filter features. As a secondary search engine, we used ACM Digital Library\footnote{\url{https://dl.acm.org/}} to challenge Scopus and enhance the literature coverage.

\textbf{Scopus} is a multidisciplinary, international database and search engine\footnote{\url{https://www.elsevier.com/products/scopus/content}}. It allows the downloading of search results in bulk and supports repeatable queries, guaranteeing reproducibility and maintainability \cite{gusenbauer_which_2020}. Scopus was highlighted in \cite{gusenbauer_which_2020,burnham_scopus_2006} as an appropriate choice due to its robust functionalities and extensive database containing more than 14,000 scientific journals. Regarding our specific area of computational linguistics, a manual search for 20 prominent conferences and journals\footnote{\url{https://scholar.google.com/citations?view_op=top_venues&hl=en&vq=eng_computationallinguistics}} confirmed that Scopus indexes all of them. The database encompasses 2689 sources (conference proceedings, journals, book series, ...) in \textit{computer science} and 364 in \textit{artificial intelligence}, highlighting its broad scope.

We used advanced searching with Scopus and the following query: 

\begin{minted}[fontsize=\scriptsize]{python}
    TITLE-ABS-KEY (
        ("Active Learning" OR "Selective Sampling") AND # End Group 1
        ("Sequence Labeling" OR "Span Categorization" OR "Entity Classification" OR 
        "Named Entity" OR "Entity Recognition" OR "Span Labeling" OR 
        "Information Extraction" OR "Sequence Tagging")) AND # End Group 2
    ( LIMIT-TO ( LANGUAGE,"English" ) ) AND 
    ( LIMIT-TO ( DOCTYPE,"cp" ) 
        OR LIMIT-TO ( DOCTYPE,"ar" ) 
        OR LIMIT-TO ( DOCTYPE,"ch" ) ) AND 
    ( LIMIT-TO ( PUBSTAGE,"final" )) # End Group 3
\end{minted}

This search query covers articles' titles, abstracts, and keywords, looking for relevant matches. It is structured with three groups of terms linked by a logical 'AND', which are indicated with the comments \textit{End Group n}. These groups include synonyms for active learning, entity recognition, and filtering criteria\footnote{We included only papers written in English that are either book chapters (ch), articles (ar) or conference papers (cp).}. 
The terms in the second group are not strict synonyms. However, in various sources, they are commonly used to describe ER. The selection of these synonyms has evolved iteratively: beginning with 'Entity Recognition', we then expanded the list by analyzing keywords in articles found through Scopus and literature surveys such as \cite{ren_survey_2022,zhang_survey_2022}, adding relevant terms gradually. 

\textbf{ACM Digital Library} serves as a secondary search engine to complement our primary database search with Scopus. Its \textit{ACM Guide to Computing Literature} database indexes over 2.8 million records\footnote{\url{https://libraries.acm.org/digital-library/acm-guide-to-computing-literature}}, emphasizing conference proceedings, a key source of current research. 

ACM does not offer to search titles, abstracts and keywords as a whole. Thus, we adapted our Scopus search string and applied it only to abstracts, leaving out our Scopus-specific filtering criteria.
We hypothesized that research papers that address AL and ER would likely include relevant keywords in their abstracts. 

\subsection{Review Process}
\label{sec:review-process}

\begin{figure}[htb]
    \centering
    \includegraphics[width=0.9\linewidth]{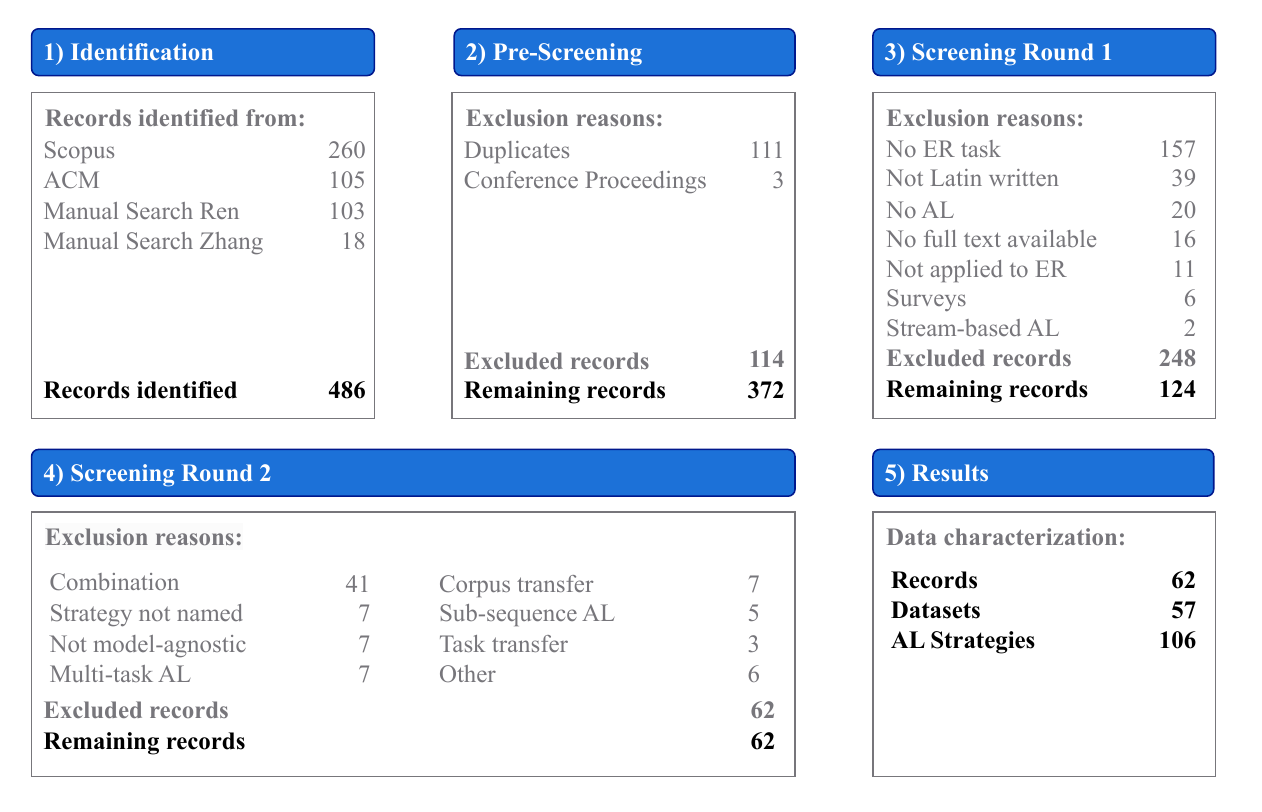}
    \caption{Our review process followed the procedure proposed by \cite{triccoPRISMAExtensionScoping2018}. It is divided into five stages, described in more detail in \autoref{sec:review-process}. The number of exclusion reasons listed for stage 2) to 4) does not always add up to the total number of excluded records because multiple exclusion criteria can exclude the same record. See the GitHub repository for a detailed list.}
    \label{fig:review-process}
\end{figure}

Figure \ref{fig:review-process} gives an overview of our review process:
First, for the \textit{Identification} of relevant papers (\textit{records}), we entered the search strings stated in \autoref{sec:search} in the advanced search fields on Scopus and ACM. We last updated the results of this search on the 12th of January, 2024. This search yielded 260 papers at Scopus and 105 papers at ACM. To further secure the comprehensiveness of our results, we manually compared the results with two AL literature surveys \cite{ren_survey_2022,zhang_survey_2022}. Together, these two surveys yielded 121 papers: 
103 papers from \cite{ren_survey_2022} and 18 papers from \cite{zhang_survey_2022}\footnote{We focused our manual search on the relevant sections to avoid including papers not targeting our topic. We extracted references from Section 3 and 4.2.4 of \cite{ren_survey_2022}. From \cite{zhang_survey_2022}, we took the sources listed in appendix A for (named) ER.}. 

We imported the results from Scopus, ACM, and the literature surveys (486 papers) to the screening and documentation tool \textit{rayyan.ai}\footnote{\url{https://www.rayyan.ai/}}. This tool facilitates documenting the results of the automatic \textit{Pre-screening} and the following manual \textit{Screening rounds 1 + 2} by recording the reviewers' decisions and exclusion reasons. During \textit{Pre-Screening}, we excluded duplicates and conference proceedings. Then, in \textit{Screening Round 1}, one reviewer screened all documents and excluded only those fulfilling at least one of the \textit{obvious exclusion criteria} (\autoref{sec:eligibility}). Thereby, the reviewer excluded 248 documents. Then, in \textit{Screening Round 2}, two reviewers screened the remaining 124 papers independently and analyzed them regarding their fit to our \textit{detailed exclusion criteria} (\autoref{sec:eligibility}). In this stage, we excluded 2 of the 3 remaining papers of our manual search that had surpassed the screening process so far. All other 118 papers had already been excluded in earlier stages: 25 were excluded during \textit{Pre-Screening}, 93 during \textit{Screening Round 1}. The resulting exclusion of more than 99\% of the manually added records indicates a high coverage of our Scopus and ACM search results.

Finally, we analyzed the remaining 62 papers and created the \textit{results} (\autoref{sec:results}): One reviewer extracted the AL strategies, the datasets, metrics, used hardware, and execution times. The second reviewer verified these results to improve the outcome's quality and coverage. Overall, we identified 106 AL strategies applied to 57 datasets. See our GitHub repository for details.

The distinction between different AL strategies in the context of ER is not trivial. The scores are often calculated at the token level, which must be aggregated to select entire documents. We consider two AL strategies as different if they differ on at least one level: e.g. if the \textit{Least Confidence (LC)} score is calculated on the token level, some authors average all token scores to a document-level LC score. Others use the value of the token with minimal confidence. This difference is represented in our analysis by identifying two separate AL strategies. 

We categorize the identified strategies according to the taxonomy already described in \autoref{sec:def-al}.

\subsection{Eligibility Criteria} 
\label{sec:eligibility}

We defined inclusion and exclusion criteria based on our \textit{review questions}  in \autoref{sec:intro}. We used these criteria to perform our scoping review. Furthermore, they help other researchers reproduce or update this scoping review. We did not apply any restrictions on the papers' publication year. All other criteria are listed below.

A paper had to match the following \textit{inclusion criteria} holistically in order to be included in the review. The paper had to:
\begin{itemize}
    \item apply a pool-based AL method as defined in \autoref{sec:def-al}.
    \item apply AL strategies to ER as defined in \autoref{sec:def-er}.
    \item be written in English to ensure it addresses the global community.
    \item be peer-reviewed, which represents a successful prior quality assessment.
    \item use at least one model-agnostic AL strategy. We want to investigate strategies that can be applied to a broad spectrum of use cases and models.
\end{itemize}

We defined two groups of exclusion criteria to structure our review process (\autoref{fig:review-process}): \textit{Obvious exclusion criteria} contain more formal and less complex decisions that one reviewer can make based on the abstract and, if necessary, an additional short screening of the full text. \textit{Detailed exclusion criteria} require a more detailed content analysis and were therefore assessed by two reviewers independently. In this case, both reviewers read the paper carefully and analyzed its contents to make a decision.

If a paper matched one of the following \textit{obvious exclusion criteria}, we excluded it from the review:

\begin{itemize}
    \item The paper was a duplicate. Duplicates could occur because we used several search strategies and included all results in the first step.
    \item It was not possible to access a full-text version with free access, IEEE or Scopus subscription.
    \item The record was a complete conference proceeding. Conference proceedings were excluded because the relevant individual papers should also be contained in our search results.
    \item The paper conducted a survey or a systematic/scoping review. We excluded them due to the same reason as conference proceedings.
    \item The paper evaluated their AL strategies only on datasets that do not follow a language based on the Latin writing system. This creates a language group with a common base, which is essential for the model selection \cite{conneau_cross-lingual_2019}.
    \item The record did not report on an ER task.
    \item The paper did not use AL.
    \item AL was not applied to an ER task.
    \item The paper used a stream-based AL procedure.
\end{itemize}

If a paper matched one of the \textit{detailed exclusion criteria}, we excluded it from our review:
\begin{itemize}
    \item Used AL strategies were not identifiable. In that case, the paper does not focus on AL as a main topic, which does not align with our objective.
    \item None of the AL strategies presented were model-agnostic.
    \item AL was applied to multiple NLP tasks in a non-separable manner.
    \item AL was combined with other methods\footnote{Data augmentation \cite{li_framework_2021}, weak \cite{gonsior_weakal_2020} and distant \cite{lee_bagging-based_2016} supervision, proactive learning \cite{li_proactive_2017},
    over-labeling \cite{mo_learning_2017}, semi supervised learning \cite{tran_hybrid_2017}, self learning \cite{neto_deep_2021}, self-training \cite{zhong_chinese_2014}, multi-task AL \cite{zhou_mtaal_2021}, pre-tagging \cite{marcheggiani_experimental_2014}, cross-lingual transfer learning  \cite{chaudhary_little_2019}, and imitation learning \cite{liu_learning_2018}} to reduce the annotation costs. We excluded the combination because these methods introduce different changes to the AL cycle (\autoref{sec:def-al}).
    \item The paper's entity recognition task did not match our definition from \autoref{sec:def-er}. Observed modifications were transfer knowledge (e.g., using a source corpus to transfer knowledge onto a target corpus \cite{lin_continuous_2020,tang_learning_2021}) or selecting subsequences instead of whole samples \cite{radmard_subsequence_2021,liu_easal_2023}.
\end{itemize}

\section{Results}
\label{sec:results}

\begin{figure}[htb]
    \centering
    \includegraphics[width=0.7\linewidth]{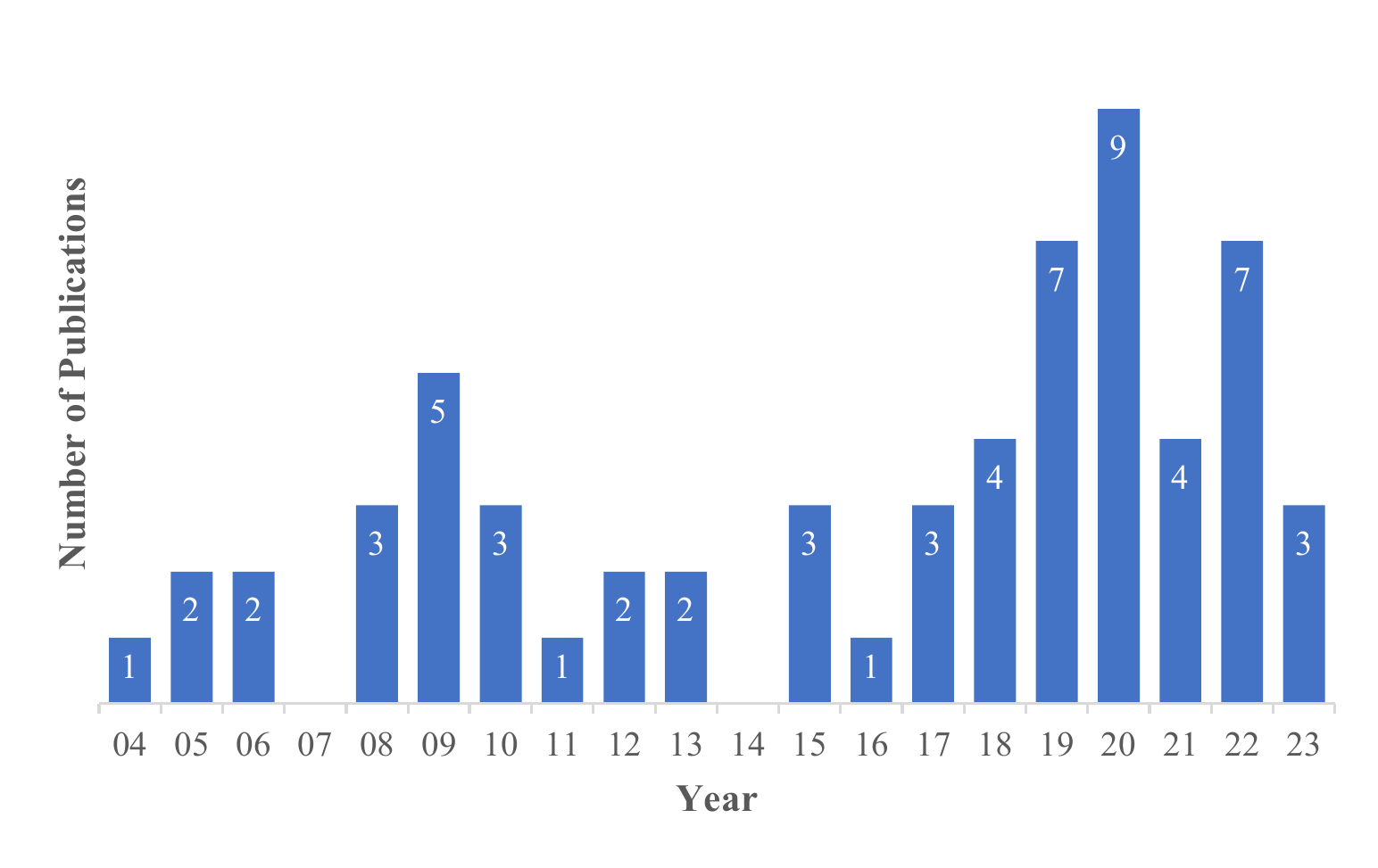}
    \caption{Publication year of the 2000er for the 62 papers analyzed within this scoping review.}
    \label{fig:years}
\end{figure}

The following sections discuss and summarize our results. For comprehensive lists of papers, AL strategies, corpora, and evaluation environments with detailed information, please consult our GitHub repository\footnote{\url{https://github.com/philipp-kohl/scoping-review-active-learning-er}}. The analyses presented in the following answer our review questions from \autoref{sec:intro}. Furthermore, we identify \textit{research gaps}, which we provide after our observations.

All papers identified through the procedure described in \autoref{sec:review-process} were published between 2004 and 2023 (compare \autoref{fig:years}). As shown in the figure, the interest in AL for ER is on the rise. Almost half of the papers (30 out of 62) were published in the past five years.

When looking at our results, our eligibility criteria must be kept in mind. We selected papers presenting research on developing or modifying AL strategies for ER. This introduces a bias that hinders the transfer of our results outside of this scientific scope.


\begin{table}[htb]
\caption{Overview of the AL strategies identified divided into categories following \cite{zhang_survey_2022,settles_analysis_2008,settles_active_2009}. More details concerning the concrete selection strategies can be found in the references and in our GitHub repository. }
\begin{tabular}{|l|l|c|c|l|}
\hline
\begin{tabular}[c]{@{}l@{}}\textbf{AL}\\\textbf{method} \end{tabular}                  & \textbf{Specification}                                                                 &\begin{tabular}[c]{@{}l@{}} \textbf{\# of AL} \\ \textbf{strategies} \end{tabular}&\begin{tabular}[c]{@{}l@{}} \textbf{\# of} \\ \textbf{usages} \end{tabular}& \textbf{Papers}                                                                                                                                                                                         \\ \hline
\multirow{4}{*}{Exploitation} & Uncertainty                                                             & \textbf{36} & \textbf{97}                   & \begin{tabular}[c]{@{}l@{}}\cite{liu_ltp_2022,li_ud_bbc_2022,mendonca_query_2020,yao_looking_2020,shrivastava_iseql_2020,agrawal_active_2021,chen_active_2017,tomanek_annotation_2010,ni_fast_2015,tomanek_proper_2009,zheng_opentag_2018}\\\cite{sapci_focusing_2023,agrawal_multicore_2023,culotta_corrective_2006,culotta_reducing_2005,zheng_opentag_2018,skeppstedt_visualising_2020,laws_active_2011,ni_fast_2015,esuli_sentence-based_2010,miller_name_2004,mejer_confidence_2010}\\
\cite{settles_analysis_2008,tang_towards_2022,shardlow_text_2019,han_clustering_2016,tchoua_active_2019,nguyen_active_2013,kim_mmr-based_2006,li_iekm-md_2020,lin_alpacatag_2019,van_nguyen_famie_2022,moniz_efficiently_2022} \\
\cite{siddhant_deep_2018,chang_using_2020,zhou_active_2023,shelmanov_active_2021,shen_deep_2017,linh_loss-based_2021,simpson_bayesian_2019,claveau_strategies_2018,saha_active_2012,shen_deep_2018,kholghi_external_2015}\\
\cite{pradhan_knowledge_2020,shen_deep_2017,verma_ensemble_2013,chen_study_2015}\end{tabular}\\ \cline{2-5} 
                              & Disagreement                                                                  & 14 &23                   & \begin{tabular}[c]{@{}l@{}}\cite{yao_looking_2020,shen_deep_2018,siddhant_deep_2018,shelmanov_active_2021,chang_using_2020,shen_deep_2017,hahn_active_2012,tomanek_reducing_2009,olsson_intrinsic_2009,settles_analysis_2008,culotta_corrective_2006}\\ \cite{olsson_privacy_2009,hachey_investigating_2005,tomanek_approximating_2008}\end{tabular}                                                                        \\ \cline{2-5} 
                              & \begin{tabular}[c]{@{}l@{}}Performance\\ Prediction \end{tabular}                                                      & 9 & 10                    & \cite{chang_using_2020,han_clustering_2016,nguyen_active_2013,linh_loss-based_2021,settles_analysis_2008}                                                                                                                                                                          \\ \cline{2-5} 
                              & Variance Reduction                                                            & 1  &1                    & \cite{settles_analysis_2008}                                                                                                                                                                                                                                                                                                                                                                                        \\ \hline
\multirow{3}{*}{Exploration}  & Density                                                                       & 6 & 6                    & \cite{chang_using_2020,chen_study_2015,han_clustering_2016,zhou_active_2023,claveau_strategies_2018,van_nguyen_famie_2022}                                                                                                                                                                          \\ \cline{2-5} 
                              & Discriminative                                                                & 5 & 6                  & \cite{li_ud_bbc_2022,chang_using_2020,chen_study_2015} \\ \cline{2-5} 
                             & \begin{tabular}[c]{@{}l@{}}Density \& \\Discriminative \end{tabular}                                                    & 3 & 3                    & \cite{mendonca_query_2020,kholghi_external_2015}                                                                                                \\ \hline
\multirow{4}{*}{Hybrid}  & \begin{tabular}[c]{@{}l@{}}Uncertainty\\ \& Density\end{tabular}        & 14 & 18                   & \cite{kim_mmr-based_2006,van_nguyen_famie_2022,linh_loss-based_2021,kholghi_clinical_2017,chen_active_2017,wei_cost-aware_2019,settles_analysis_2008,mendonca_query_2020,claveau_strategies_2018,zhou_active_2023,veerasekharreddy_named_2022}  \\ \cline{2-5}  
                              & \begin{tabular}[c]{@{}l@{}}Uncertainty\\ \& Discriminative\end{tabular} & 12 & 13                   & \cite{shen_deep_2018,chang_using_2020,kholghi_clinical_2017,kholghi_external_2015,sapci_focusing_2023}                                                                                                                                                                                                                                                                                               \\ \cline{2-5} 
                              & \begin{tabular}[c]{@{}l@{}}Uncertainty\\ \& Other\end{tabular}          & 5 & 5                    & \cite{kholghi_external_2015,tchoua_active_2019,brent_systematic_2009}                                                                                                                                               \\ \cline{2-5} 
                              & \begin{tabular}[c]{@{}l@{}}Disagreement \&\\ Discriminative\end{tabular}      & 1  & 1                    & \cite{gao_active_2019}                                                                                                                                                                                                                                                                                       \\ \hline
\end{tabular}
\label{table:strategies}
\end{table}


\begin{table}[tb]
  \centering
  \caption{Uncertainty-based AL strategies with their heuristic.}
  \label{table:uncertainty_heuristics}
  \begin{tabular}{l|c|c}
    \hline
    \textbf{Heuristic} & \textbf{\# of strategies} & \textbf{\# of usages} \\
    \hline
    Least Confidence  & 11 & 35 \\
    Entropy        & 9  & 22 \\
    Margin          & 4  & 14 \\
    Count           & 4  & 4 \\
    Round Robin     & 3  & 3 \\
    Max. Norm. Log-Probability & 1 & 15 \\
    Other           & 4  & 4 \\
    \hline
    \textbf{Sum}      & 36 & 97 \\
    \hline
  \end{tabular}
\end{table}

\subsection{Active Learning Strategies for ER (Review Question 1)}
\label{sec:al-strats}
In total, we identified 106 AL strategies with ER applications in our scoping review. Table \ref{table:strategies} lists the total number of AL strategies using the different methods (exploitation, exploration, and hybrid) and their specification following \cite{zhang_survey_2022,settles_analysis_2008,settles_active_2009}.

We list our observations of the results regarding the AL strategies in the following: 

\begin{description}
    \item[Focus on Exploitation-based Approaches] 
    ~ We identified 60  exploitation-based AL strategies, which were used 131 times in our 62 analyzed papers. Un\-certainty-based AL strategies represent the majority (\autoref{table:strategies}): 60\% of the exploitation-based strategies belong to uncertainty-based approaches, which are used 74\% of the time. \autoref{table:uncertainty_heuristics} shows the number of strategies and their usages of the uncertainty-based AL strategies grouped by the different scoring approaches, which we call \textit{heuristics}.
    \textit{Least confidence} approaches were developed and used most. 

    \item[Infrequently Used Exploration-based Approaches] ~ We identified 14 ex\-ploration-based strategies. They are applied less often in isolation (15 times) than in hybrid settings (37 times). This is noteworthy because the implementation of hybrid approaches is more complex. Exploitation-based strategies, in contrast, are used extensively on their own.

    \item[Distribution of AL Strategies Over Domains] As depicted in \autoref{table:domain_strategies}, the three most used domains are bio-medicine, medicine, and newspaper. We observed that significantly more exploitation than exploration approaches are applied in all domains. Interestingly, hybrid strategies are on par with exploitation strategies in the medical domain. The other two domains use less than half as often hybrid strategies as exploitation strategies.
\end{description}

\begin{table}[htb]
  \centering
  \caption{Number of strategies applied to the three main domains.}
  \label{table:domain_strategies}
  \begin{tabular}{l|c|c|c|c}
    \hline
    \textbf{Approach} & \textbf{\# of strategies} & \textbf{Medicine} & \textbf{Biomedicine} & \textbf{News-corpora} \\
    \hline
    Exploitation  & 60 & 17 & 29 & 44 \\
    Exploration  & 14 & 5  & 3  & 4 \\
    Hybrid       & 32 & 17 & 13 & 14 \\
    \hline
    \textbf{Sum} & 106 & 39 & 45 & 62 \\
    \hline
  \end{tabular}
\end{table}

\subsection{Corpora (Review Question 2a)}

We identified 57 corpora from more than 9 domains. Most corpora belong to the domain of bio-medicine (12), medicine (9), and newspaper articles (7). The others hold three or fewer corpora\footnote{Cybersecurity (3), Scientific Papers (3), Twitter Posts (2), Wikipedia Articles (3), Instructions (2), E-Mail (2), and we group the single domain corpora into an \textit{other} category (14). See GitHub repository for an exhaustive list.}. 

\begin{figure}[htb]
    \centering   \includegraphics[width=0.85\linewidth]{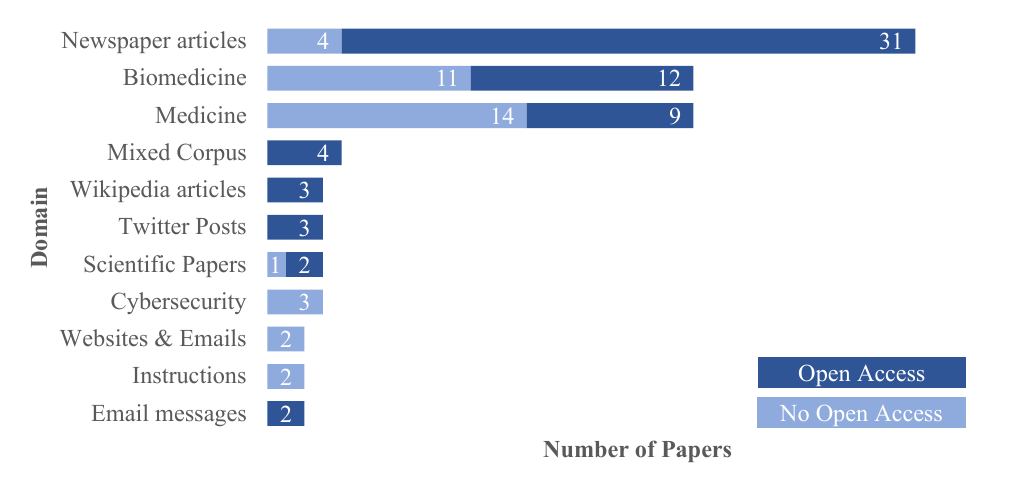}
    \caption{The figure shows how many times corpora from specific domains are used in our 62 papers, grouped by the corpus licensing. The majority of the papers use open access corpora for their experiments.}
    \label{fig:corpora-usage}
\end{figure}

\autoref{fig:corpora-usage} shows the corpora usage per domain. 35 out of 62 papers use newspaper articles to investigate AL performance. Second and third are bio-medicine and medicine, with 23 usages each. 
The bio-medicine and medicine domains have the highest number of corpora, but researchers use newspaper articles more frequently. 

The CoNLL \cite{tjong_kim_sang_introduction_2003} corpora (2003 and 2002) based on newspaper articles were the most used corpora with 30 usages. Other often used corpora were i2b2/VA 2010 (medicine) \cite{uzuner_2010_2011} with 7 usages and JNLPBA (bio-medicine) \cite{collier_introduction_2004} with 6 usages.

We investigated public access to corpora and prepared a list of accessible datasets for further research. We consider a corpus \textit{open access} if the researchers provide a link to the dataset or to a reference that introduced and published the corpus. Also, we consulted the authors' web pages to find corpora for their publications when necessary. If a corpus has to be requested and is only licensable for research and academic purposes, we consider it as open access.
26 out of 57 corpora follow the reproducible research requirement to publish their datasets with open access. 

Researchers do not always publish their datasets or make their annotations freely available. 26 of 57 corpora are private or can be licensed against a fee. We classify them as \textit{not open access}. For 5 corpora, we could not draw a reasonable decision due to invalid or moved internet resources. Thus, we consider them as not openly accessible.

We list our observations of the results regarding the corpora in the following. 
\begin{description}
    \item[High Usage of Newspaper Articles] Newspaper articles show the highest usage (\autoref{fig:corpora-usage}) across the domains. More than half of the identified papers evaluate their active learning strategies on newspaper articles, although the medicine and bio-medicine corpora have more corpora. We hypothesize that newspaper articles are often freely available, and the annotation process does not need the same level of expertise as for (bio-)medicine.
    \item[Most Corpora for Bio-medicine and Medicine Domain] The annotation process for the bio-medicine and medicine corpora can be very costly due to highly educated staff. 
    7 out of 21 bio-medicine and medicine corpora are accessible under open-access licensing (see GitHub).
    We hypothesize that the number of corpora indicates that the domain sees great potential to reduce the annotation effort with AL. With their contribution of publicly available corpora, which were annotated by highly educated staff, they probably want to facilitate more research. 
\end{description}

We provide a list of publicly accessible corpora designated for ER, enabling researchers to investigate and advance the development of ER and AL methods.

\subsection{Metrics, Hardware, and Execution Times (Review Questions 2b, 2c)}

Our examination reveals a uniformity in the metrics used across studies, suggesting a consensus on their effectiveness. Regarding the hardware, only 6 out of 62 papers detail the hardware used for experiments, which we consider a critical oversight given that the hardware can significantly influence the training and inference times. This impacts the annotation process: The time for retraining models and determining new data points for the annotators results in waiting times \cite{van_nguyen_famie_2022}.
The hardware reported ranged from personal computers to small server instances and workstations. No information was found on the usage of distributed clusters.
Our analysis of execution times identified 13 papers reporting on training duration, annotator wait times, inference speeds, and annotation time.

We made the following observations:
\begin{description}
    \item[AL Performance Comparison Metrics]
    Due to the differences in the implementations, parameters, and environments, a direct comparison of the performance of different AL strategies is unrealizable.
    However, the findings of the records offer information about the metrics used to evaluate AL strategies. Frequently it is F1-score (60), precision (16), and recall (16). Rarely ~ ($<$ 4 times), it is accuracy, annotation time, and error rate.  
    \item[AL Execution Times and Used Hardware] Little attention is paid to these aspects. 13 out of 62 papers reported any kind of timing information. Only 6 stated their used hardware. Papers presenting real-world applications of AL to ER tasks \cite{van_nguyen_famie_2022} mention the relevance of short retraining times for the AL model because they correlate with the waiting time for annotators. We could not find information about the duration of the initialization of an AL strategy, nor did we find information about the time a strategy needs to propose the new data points to the annotator. 
\end{description}

\subsection{Research Gaps}

Based on our observations from the last sections, we formulate open topics as questions, which can guide future work in the field of AL and ER:
\begin{description}
    \item[General] How do AL strategies perform in various domains in terms of their performance and execution time on specified hardware conditions? Is there a universally effective active learning strategy independent of the use case?
    \item[Exploitation Approaches] What are the reasons for the quantitative dominance of exploitation-based, especially uncertainty-based, AL strategies? Are these strategies also outstanding qualitatively? Are they used as a solution for ER, or are they primarily used as baselines for comparison with other strategies?
    \item[Exploration and Hybrid Approaches] Is the isolated usage of exploration-based AL strategies less beneficial than solely exploitation-based applications? Do hybrid approaches outperform exploration but not exploitation-based strategies?
    \item[Domain Research] Why does medical research for ER focus equally on exploitation and hybrid approaches while other domains favor exploitation-based strategies?
    Does the intense focus on evaluating AL strategies on newspaper articles reveal well-performing strategies? Do newspaper articles function as a baseline?
    Do AL strategies perform on specialized domains such as medicine as well as for broad domains like newspaper articles? Do the corpora and label sets differ in number and complexity?
    Is there a universally effective active learning strategy, or are certain strategies more effective in specific domains?
\end{description}

\subsection{Evaluation Environment}
According to the results presented, we establish a set of criteria for evaluating the effectiveness of Active Learning (AL) strategies in future works. An evaluation framework like ALE \cite{kohl_ale_2023} and a reasoned selection of AL strategies and datasets enable a fair comparison. The evaluation environment should consider the following aspects:

\begin{description}
    \item[Strategies] The comparison of strategies should include at least one strategy of each specification and heuristic (see \autoref{table:strategies} and \autoref{table:uncertainty_heuristics}). To assess the impact of the aggregation method, the permutation should be considered for exploitation and hybrid strategies. 
    \item[Dataset] The strategies should be tested on a diverse set of corpora. This assesses the strategy's robustness and allows to investigate the existence of an overall high-performing AL strategy. Based on the open access corpora (\autoref{fig:corpora-usage}), several domains can be tested (news, (bio-) medicine, scientific, and social media). The corpora may differ in size, language, and label complexity, which introduces different challenges.
    \item[Hardware and Execution Time] It is important to consider the time constraints of AL strategies as they can affect the annotation process \cite{herde_survey_2021}: The time required for proposing new data points and retraining the model can impact the waiting time for annotators. Therefore, it is essential to record the timings for initializing the AL strategy, proposing data points, and retraining the model. The timing information is strongly dependent on the hardware used. Therefore, it is crucial to provide information about the used hardware.
    \item[Evaluation metric] Based on this scoping review, the F1-Score should be in the list of reporting metrics.
    \item[Bias] Tracking bias reinforcement can help identify strategies that mitigate bias instead of amplifying it (see \autoref{sec:ethical}).
\end{description}

\section{Related Work}
\label{sec:related-work}
Besides AL, researchers developed other approaches to reduce the annotation effort.
Semi-supervised \cite{sintayehu_named_2021} and weak supervision \cite{lison_skweak_2021} techniques rely on an initial dataset from which they derive rules or heuristics, enabling the automatic annotation of a larger dataset with reduced manual effort. These approaches might introduce noise into the data due to less precise heuristics.

Distant supervision \cite{hedderich_anea_2021}, on the other hand, leverages external resources to generate positive examples for specific tasks, which is especially useful when external knowledge bases can provide substantial input. Data augmentation \cite{feng_survey_2021} complements these methods by creating new instances from already labeled data through various linguistic transformations.

AL and the stated methods can be used together to further reduce the manual effort  \cite{tran_hybrid_2017,gonsior_weakal_2020,lee_bagging-based_2016,li_framework_2021}.

Surveys such as those by \cite{ren_survey_2022,zhang_survey_2022} are most closely related to our work. Thereby, \cite{ren_survey_2022} considers deep learning techniques with AL in several areas (such as computer vision), while \cite{zhang_survey_2022} focuses solely on NLP. Both surveys categorize the strategies found, many of which cannot be directly applied to ER.

\section{Ethical Consideration}
\label{sec:ethical}

AL must be ethically scrutinized in the general context of NLP \cite{leidner_ethical_2017}. Specifically, AL as a data selection method can insert or enforce statistical bias following \cite{farquhar_statistical_2021,hassan_d-calm_nodate}. The authors propose possible reasons for this unwanted effect, such as the distribution of data points in the seed and train dataset. Furthermore, AL errors can cause the model to become wrongly confident, making it difficult to correct the learned structure. As a result, data may stop being proposed for this concept because the model does not show room for improvement \cite{hassan_d-calm_nodate}. Another bias may be transferred from transformer models \cite{vaswani_attention_2023} during the pre-training phase of the AL model \cite{hassan_d-calm_nodate}.

These bias-introducing and enforcing effects are especially alarming considering the focus of AL research on the medical domain.  
Existing approaches optimize for fairness metrics \cite{anahideh_fair_2021} or vary the error rate in each iteration of adaptive clustering to reduce bias \cite{hassan_d-calm_nodate} for classification tasks. These methods should also be tested for ER.

\section{Conclusion and Future Work}
\label{sec:conclusion}
We conducted a scoping review to provide an overview of active learning strategies, metrics, datasets, execution times, and hardware used for the entity recognition task. 

Our results as comprehensive lists can be found in the provided GitHub repository: We identified 106 AL strategies and 57 datasets in 62 papers. A large share of the strategies follows the exploitation-based (60) approach. 36 of them use uncertainty-based sampling. Furthermore, we noted fewer exploration-based AL strategies than hybrid ones. For evaluation purposes, the F1-score is the dominant metric to demonstrate the performance of an AL strategy. Unfortunately, very few researchers report the execution time and used hardware for their experiments. We examined the 57 datasets on their availability and found 26 publicly accessible corpora. The most frequently used corpora are from the newspaper, bio-medical, and medical domains. 
We created an evaluation environment based on our observations. 
Additionally, we have identified research gaps in the field, which researchers can use as an outline for further research.

We plan to conduct comprehensive performance tests on a subset of the AL strategies and datasets found in this scoping review based on our evaluation environment.

%
%
%
\bibliographystyle{splncs04}
\bibliography{delta2024,algroup,alspanlabeling,relatedwork}

\begin{thebibliography}{100}
\providecommand{\url}[1]{\texttt{#1}}
\providecommand{\urlprefix}{URL }
\providecommand{\doi}[1]{https://doi.org/#1}

\bibitem{agrawal_active_2021}
Agrawal, A., Tripathi, S., Vardhan, M.: Active learning approach using a modified least confidence sampling strategy for named entity recognition. Progress in Artificial Intelligence  \textbf{10}(2),  113--128 (2021). \doi{10.1007/s13748-021-00230-w}

\bibitem{agrawal_multicore_2023}
Agrawal, A., Tripathi, S., Vardhan, M.: Multicore based least confidence query sampling strategy to speed up active learning approach for named entity recognition. Computing  \textbf{105}(5),  979--997 (2023). \doi{10.1007/s00607-021-01000-1}

\bibitem{anahideh_fair_2021}
Anahideh, H., Asudeh, A., Thirumuruganathan, S.: Fair {{Active Learning}}. arXiv:2001.01796 [cs, stat]  (Mar 2021)

\bibitem{arora_active_2007}
Arora, S., Agarwal, S., Students, M.: Active {{Learning}} for {{Natural Language Processing}}. Language Technologies Institute School of Computer Science Carnegie Mellon University  \textbf{2} (2007)

\bibitem{bondu_exploration_2010}
Bondu, A., Lemaire, V., Boull{\'e}, M.: Exploration vs. exploitation in active learning : {{A Bayesian}} approach. In: The 2010 {{International Joint Conference}} on {{Neural Networks}} ({{IJCNN}}). pp.~1--7 (Jul 2010). \doi{10.1109/IJCNN.2010.5596815}

\bibitem{brent_systematic_2009}
Brent, P., Green, N., Breimyer, P., Krishnamurthy, R., Samatova, N.: Systematic evaluation of convergence criteria in iterative training for {{NLP}}. In: Proceedings of the 22nd {{International Florida Artificial Intelligence Research Society Conference}}, {{FLAIRS-22}}. pp. 15--20 (2009)

\bibitem{brownLanguageModelsAre2020}
Brown, T.B., Mann, B., Ryder, N., Subbiah, M., Kaplan, J., Dhariwal, P., Neelakantan, A., Shyam, P., Sastry, G., Askell, A., Agarwal, S., {Herbert-Voss}, A., Krueger, G., Henighan, T., Child, R., Ramesh, A., Ziegler, D.M., Wu, J., Winter, C., Hesse, C., Chen, M., Sigler, E., Litwin, M., Gray, S., Chess, B., Clark, J., Berner, C., McCandlish, S., Radford, A., Sutskever, I., Amodei, D.: Language {{Models}} are {{Few-Shot Learners}} (Jul 2020). \doi{10.48550/arXiv.2005.14165}

\bibitem{brown_language_2020}
Brown, T.B., Mann, B., Ryder, N., Subbiah, M., Kaplan, J., Dhariwal, P., Neelakantan, A., Shyam, P., Sastry, G., Askell, A., Agarwal, S., {Herbert-Voss}, A., Krueger, G., Henighan, T., Child, R., Ramesh, A., Ziegler, D.M., Wu, J., Winter, C., Hesse, C., Chen, M., Sigler, E., Litwin, M., Gray, S., Chess, B., Clark, J., Berner, C., McCandlish, S., Radford, A., Sutskever, I., Amodei, D.: Language models are few-shot learners. In: Proceedings of the 34th {{International Conference}} on {{Neural Information Processing Systems}}. pp. 1877--1901. {{NIPS}}'20, {Curran Associates Inc.}, {Red Hook, NY, USA} (Dec 2020)

\bibitem{burnham_scopus_2006}
Burnham, J.F.: Scopus database: A review. Biomedical Digital Libraries  \textbf{3}(1), ~1--8 (Dec 2006). \doi{10.1186/1742-5581-3-1}

\bibitem{chang_using_2020}
Chang, H.S., Vembu, S., Mohan, S., Uppaal, R., McCallum, A.: Using error decay prediction to overcome practical issues of deep active learning for named entity recognition. Machine Learning  \textbf{109}(9-10),  1749--1778 (2020). \doi{10.1007/s10994-020-05897-1}

\bibitem{chaudhary_little_2019}
Chaudhary, A., Xie, J., Sheikh, Z., Neubig, G., Carbonell, J.: A little annotation does a lot of good: {{A}} study in bootstrapping low-resource named entity recognizers. In: {{EMNLP-IJCNLP}} 2019 - 2019 {{Conference}} on {{Empirical Methods}} in {{Natural Language Processing}} and 9th {{International Joint Conference}} on {{Natural Language Processing}}, {{Proceedings}} of the {{Conference}}. pp. 5164--5174 (2019)

\bibitem{chen_active_2017}
Chen, Y., Lask, T., Mei, Q., Chen, Q., Moon, S., Wang, J., Nguyen, K., Dawodu, T., Cohen, T., Denny, J., Xu, H.: An active learning-enabled annotation system for clinical named entity recognition. BMC Medical Informatics and Decision Making  \textbf{17} (2017). \doi{10.1186/s12911-017-0466-9}

\bibitem{chen_study_2015}
Chen, Y., Lasko, T.A., Mei, Q., Denny, J.C., Xu, H.: A study of active learning methods for named entity recognition in clinical text. Journal of Biomedical Informatics  \textbf{58},  11--18 (Dec 2015). \doi{10.1016/j.jbi.2015.09.010}

\bibitem{claveau_strategies_2018}
Claveau, V., Kijak, E.: Strategies to select examples for active learning with conditional random fields. In: Lecture {{Notes}} in {{Computer Science}} (Including Subseries {{Lecture Notes}} in {{Artificial Intelligence}} and {{Lecture Notes}} in {{Bioinformatics}}). vol. 10761 LNCS, pp. 30--43 (2018). \doi{10.1007/978-3-319-77113-7\_3}

\bibitem{collier_introduction_2004}
Collier, N., Ohta, T., Tsuruoka, Y., Tateisi, Y., Kim, J.D.: Introduction to the {{Bio-entity Recognition Task}} at {{JNLPBA}}. In: Collier, N., Ruch, P., Nazarenko, A. (eds.) Proceedings of the {{International Joint Workshop}} on {{Natural Language Processing}} in {{Biomedicine}} and Its {{Applications}} ({{NLPBA}}/{{BioNLP}}). pp. 73--78. {COLING}, {Geneva, Switzerland} (Aug 2004)

\bibitem{conneau_cross-lingual_2019}
Conneau, A., Lample, G.: Cross-lingual language model pretraining. In: Proceedings of the 33rd {{International Conference}} on {{Neural Information Processing Systems}}, pp. 7059--7069. No.~634, {Curran Associates Inc.}, {Red Hook, NY, USA} (Dec 2019)

\bibitem{culotta_corrective_2006}
Culotta, A., Kristjansson, T., McCallum, A., Viola, P.: Corrective feedback and persistent learning for information extraction. Artificial Intelligence  \textbf{170}(14-15),  1101--1122 (2006). \doi{10.1016/j.artint.2006.08.001}

\bibitem{culotta_reducing_2005}
Culotta, A., McCallum, A.: Reducing labeling effort for structured prediction tasks. In: Proceedings of the {{National Conference}} on {{Artificial Intelligence}}. vol.~2, pp. 746--751 (2005)

\bibitem{devlinBERTPretrainingDeep2018}
Devlin, J., Chang, M.W., Lee, K., Toutanova, K.: {{BERT}}: {{Pre-training}} of {{Deep Bidirectional Transformers}} for {{Language Understanding}}. https://arxiv.org/abs/1810.04805v2 (Oct 2018)

\bibitem{esuli_sentence-based_2010}
Esuli, A., Marcheggiani, D., Sebastiani, F.: Sentence-based active learning strategies for information extraction. In: {{CEUR Workshop Proceedings}}. vol.~560, pp. 41--45 (2010)

\bibitem{farquhar_statistical_2021}
Farquhar, S., Gal, Y., Rainforth, T.: On {{Statistical Bias In Active Learning}}: {{How}} and {{When To Fix It}} (May 2021)

\bibitem{feng_survey_2021}
Feng, S.Y., Gangal, V., Wei, J., Chandar, S., Vosoughi, S., Mitamura, T., Hovy, E.: A {{Survey}} of {{Data Augmentation Approaches}} for {{NLP}} (Dec 2021). \doi{10.48550/arXiv.2105.03075}

\bibitem{gao_active_2019}
Gao, N., Karampatziakis, N., Potharaju, R., Cucerzan, S.: Active entity recognition in low resource settings. In: International {{Conference}} on {{Information}} and {{Knowledge Management}}, {{Proceedings}}. pp. 2261--2264 (2019). \doi{10.1145/3357384.3358109}

\bibitem{gonsior_weakal_2020}
Gonsior, J., Thiele, M., Lehner, W.: {{WeakAL}}: {{Combining Active Learning}} and {{Weak Supervision}}. In: Lecture {{Notes}} in {{Computer Science}} (Including Subseries {{Lecture Notes}} in {{Artificial Intelligence}} and {{Lecture Notes}} in {{Bioinformatics}}). vol. 12323 LNAI, pp. 34--49 (2020). \doi{10.1007/978-3-030-61527-7\_3}

\bibitem{grant_typology_2009}
Grant, M.J., Booth, A.: A typology of reviews: An analysis of 14 review types and associated methodologies. Health Information \& Libraries Journal  \textbf{26}(2),  91--108 (2009). \doi{10.1111/j.1471-1842.2009.00848.x}

\bibitem{gusenbauer_which_2020}
Gusenbauer, M., Haddaway, N.R.: Which academic search systems are suitable for systematic reviews or meta-analyses? {{Evaluating}} retrieval qualities of {{Google Scholar}}, {{PubMed}}, and 26 other resources. Research Synthesis Methods  \textbf{11}(2),  181--217 (2020). \doi{10.1002/jrsm.1378}

\bibitem{hachey_investigating_2005}
Hachey, B., Alex, B., Becker, M.: Investigating the effects of selective sampling on the annotation task. In: {{CoNLL}} 2005 - {{Proceedings}} of the {{Ninth Conference}} on {{Computational Natural Language Learning}}. pp. 144--151 (2005). \doi{10.3115/1706543.1706569}

\bibitem{hahn_active_2012}
Hahn, U., Beisswanger, E., Buyko, E., Faessler, E.: Active {{Learning-based}} corpus annotation--the {{PathoJen}} experience. AMIA ... Annual Symposium proceedings / AMIA Symposium. AMIA Symposium  \textbf{2012},  301--310 (2012)

\bibitem{han_clustering_2016}
Han, X., Kwoh, C., Kim, J.J.: Clustering based active learning for biomedical {{Named Entity Recognition}}. In: Proceedings of the {{International Joint Conference}} on {{Neural Networks}}. vol. 2016-October, pp. 1253--1260 (2016). \doi{10.1109/IJCNN.2016.7727341}

\bibitem{hassan_d-calm_nodate}
Hassan, S., Alikhani, M.: D-{{CALM}}: {{A Dynamic Clustering-based Active Learning Approach}} for {{Mitigating Bias}}

\bibitem{hassanzadeh_two-phase_2013}
Hassanzadeh, H., Keyvanpour, M.: A two-phase hybrid of semi-supervised and active learning approach for sequence labeling. Intelligent Data Analysis  \textbf{17}(2),  251--270 (2013). \doi{10.3233/IDA-130577}

\bibitem{hedderich_anea_2021}
Hedderich, M.A., Lange, L., Klakow, D.: {{ANEA}}: {{Distant Supervision}} for {{Low-Resource Named Entity Recognition}} (Apr 2021). \doi{10.48550/arXiv.2102.13129}

\bibitem{herde_survey_2021}
Herde, M., Huseljic, D., Sick, B., Calma, A.: A {{Survey}} on {{Cost Types}}, {{Interaction Schemes}}, and {{Annotator Performance Models}} in {{Selection Algorithms}} for {{Active Learning}} in {{Classification}}. IEEE Access  \textbf{9},  166970--166989 (2021). \doi{10.1109/ACCESS.2021.3135514}

\bibitem{jayakumar_large_2023}
Jayakumar, T., Farooqui, F., Farooqui, L.: Large {{Language Models}} are legal but they are not: {{Making}} the case for a powerful {{LegalLLM}}. In: {Preo{\textbackslash}textcommabelowtiuc-Pietro}, D., Goanta, C., Chalkidis, I., Barrett, L., Spanakis, G.J., Aletras, N. (eds.) Proceedings of the {{Natural Legal Language Processing Workshop}} 2023. pp. 223--229. {Association for Computational Linguistics}, {Singapore} (Dec 2023). \doi{10.18653/v1/2023.nllp-1.22}

\bibitem{kholghi_clinical_2017}
Kholghi, M., De~Vine, L., Sitbon, L., Zuccon, G., Nguyen, A.: Clinical information extraction using small data: {{An}} active learning approach based on sequence representations and word embeddings. Journal of the Association for Information Science and Technology  \textbf{68}(11),  2543--2556 (2017). \doi{10.1002/asi.23936}

\bibitem{kholghi_external_2015}
Kholghi, M., Sitbon, L., Zuccon, G., Nguyen, A.: External knowledge and query strategies in active learning: A study in clinical information extraction. In: International {{Conference}} on {{Information}} and {{Knowledge Management}}, {{Proceedings}}. vol. 19-23-Oct-2015, pp. 143--152 (2015). \doi{10.1145/2806416.2806550}

\bibitem{kim_mmr-based_2006}
Kim, S., Song, Y., Kim, K., Cha, J.W., Lee, G.: {{MMR-based}} active machine learning for bio named entity recognition. In: {{HLT-NAACL}} 2006 - {{Human Language Technology Conference}} of the {{North American Chapter}} of the {{Association}} of {{Computational Linguistics}}, {{Short Papers}}. pp. 69--72 (2006)

\bibitem{kohl_ale_2023}
Kohl, P., Freyer, N., Kr{\"a}mer, Y., Werth, H., Wolf, S., Kraft, B., Meinecke, M., Z{\"u}ndorf, A.: {{ALE}}: {{A Simulation-Based Active Learning Evaluation Framework}} for the {{Parameter-Driven Comparison}} of {{Query Strategies}} for {{NLP}}. In: Conte, D., Fred, A., Gusikhin, O., Sansone, C. (eds.) Deep {{Learning Theory}} and {{Applications}}. pp. 235--253. Communications in {{Computer}} and {{Information Science}}, {Springer Nature Switzerland}, {Cham} (2023). \doi{10.1007/978-3-031-39059-3\_16}

\bibitem{laws_active_2011}
Laws, F., Scheible, C., Sch{\"u}tze, H.: Active learning with amazon mechanical turk. In: {{EMNLP}} 2011 - {{Conference}} on {{Empirical Methods}} in {{Natural Language Processing}}, {{Proceedings}} of the {{Conference}}. pp. 1546--1556 (2011)

\bibitem{lee_bagging-based_2016}
Lee, S., Song, Y., Choi, M., Kim, H.: Bagging-based active learning model for named entity recognition with distant supervision. In: 2016 {{International Conference}} on {{Big Data}} and {{Smart Computing}}, {{BigComp}} 2016. pp. 321--324 (2016). \doi{10.1109/BIGCOMP.2016.7425938}

\bibitem{leidner_ethical_2017}
Leidner, J.L., Plachouras, V.: Ethical by {{Design}}: {{Ethics Best Practices}} for {{Natural Language Processing}}. In: Proceedings of the {{First ACL Workshop}} on {{Ethics}} in {{Natural Language Processing}}. pp. 30--40. {Association for Computational Linguistics}, {Valencia, Spain} (2017). \doi{10.18653/v1/W17-1604}

\bibitem{li_proactive_2017}
Li, M., Nguyen, N., Ananiadou, S.: Proactive {{Learning}} for {{Named Entity Recognition}}. In: {{BioNLP}} 2017 - {{SIGBioMed Workshop}} on {{Biomedical Natural Language Processing}}, {{Proceedings}} of the 16th {{BioNLP Workshop}}. pp. 117--125 (2017)

\bibitem{li_framework_2021}
Li, Q., Huang, Z., Dou, Y., Zhang, Z.: A {{Framework}} of {{Data Augmentation While Active Learning}} for {{Chinese Named Entity Recognition}}. In: Lecture {{Notes}} in {{Computer Science}} (Including Subseries {{Lecture Notes}} in {{Artificial Intelligence}} and {{Lecture Notes}} in {{Bioinformatics}}). vol. 12816 LNAI, pp. 88--100 (2021). \doi{10.1007/978-3-030-82147-0\_8}

\bibitem{li_ud_bbc_2022}
Li, W., Du, Y., Li, X., Chen, X., Xie, C., Li, H., Li, X.: {{UD}}\_{{BBC}}: {{Named}} entity recognition in social network combined {{BERT-BiLSTM-CRF}} with active learning. Engineering Applications of Artificial Intelligence  (2022). \doi{10.1016/j.engappai.2022.105460}

\bibitem{li_iekm-md_2020}
Li, Y., Yue, T., Zhenxin, W.: {{IEKM-MD}}: {{An}} intelligent platform for information extraction and knowledge mining in multi-domains. In: {{CEUR Workshop Proceedings}}. vol.~2658, pp. 73--78 (2020)

\bibitem{lin_alpacatag_2019}
Lin, B., Lee, D.H., Xu, F., Lan, O., Ren, X.: {{AlpacaTag}}: {{An}} active learning-based crowd annotation framework for sequence tagging. In: {{ACL}} 2019 - 57th {{Annual Meeting}} of the {{Association}} for {{Computational Linguistics}}, {{Proceedings}} of {{System Demonstrations}}. pp. 58--63 (2019)

\bibitem{lin_continuous_2020}
Lin, S., Gao, J., Zhang, S., He, X., Sheng, Y., Chen, J.: A continuous learning method for recognizing named entities by integrating domain contextual relevance measurement and {{Web}} farming mode of {{Web}} intelligence. World Wide Web  \textbf{23}(3),  1769--1790 (2020). \doi{10.1007/s11280-019-00758-x}

\bibitem{linh_loss-based_2021}
Linh, L., Nguyen, M.T., Zuccon, G., Demartini, G.: Loss-based {{Active Learning}} for {{Named Entity Recognition}}. In: Proceedings of the {{International Joint Conference}} on {{Neural Networks}}. vol. 2021-July (2021). \doi{10.1109/IJCNN52387.2021.9533675}

\bibitem{lison_skweak_2021}
Lison, P., Barnes, J., Hubin, A.: Skweak: {{Weak Supervision Made Easy}} for {{NLP}}. In: Proceedings of the 59th {{Annual Meeting}} of the {{Association}} for {{Computational Linguistics}} and the 11th {{International Joint Conference}} on {{Natural Language Processing}}: {{System Demonstrations}}. pp. 337--346 (2021). \doi{10.18653/v1/2021.acl-demo.40}

\bibitem{liu_learning_2018}
Liu, M., Buntine, W., Haffari, G.: Learning how to actively learn: {{A}} deep imitation learning approach. In: {{ACL}} 2018 - 56th {{Annual Meeting}} of the {{Association}} for {{Computational Linguistics}}, {{Proceedings}} of the {{Conference}} ({{Long Papers}}). vol.~1, pp. 1874--1883 (2018). \doi{10.18653/v1/p18-1174}

\bibitem{liu_ltp_2022}
Liu, M., Tu, Z., Zhang, T., Su, T., Xu, X., Wang, Z.: {{LTP}}: {{A New Active Learning Strategy}} for {{CRF-Based Named Entity Recognition}}. Neural Processing Letters  \textbf{54}(3),  2433--2454 (2022). \doi{10.1007/s11063-021-10737-x}

\bibitem{liu_easal_2023}
Liu, Y., Hu, J., Chen, Z., Wan, X., Chang, T.H.: {{EASAL}}: {{Entity-Aware Subsequence-Based Active Learning}} for {{Named Entity Recognition}}. In: Proceedings of the 37th {{AAAI Conference}} on {{Artificial Intelligence}}, {{AAAI}} 2023. vol.~37, pp. 8897--8905 (2023)

\bibitem{loy_stream-based_2012}
Loy, C.C., Hospedales, T.M., {Tao Xiang}, {Shaogang Gong}: Stream-based joint exploration-exploitation active learning. 2012 IEEE Conference on Computer Vision and Pattern Recognition pp. 1560--1567 (Jun 2012). \doi{10.1109/CVPR.2012.6247847}

\bibitem{marcheggiani_experimental_2014}
Marcheggiani, D., Arti{\`e}res, T.: An experimental comparison of active learning strategies for partially labeled sequences. In: {{EMNLP}} 2014 - 2014 {{Conference}} on {{Empirical Methods}} in {{Natural Language Processing}}, {{Proceedings}} of the {{Conference}}. pp. 898--906 (2014). \doi{10.3115/v1/d14-1097}

\bibitem{mejer_confidence_2010}
Mejer, A., Crammer, K.: Confidence in structured-prediction using {{Confidence-Weighted}} models. In: {{EMNLP}} 2010 - {{Conference}} on {{Empirical Methods}} in {{Natural Language Processing}}, {{Proceedings}} of the {{Conference}}. pp. 971--981 (2010)

\bibitem{mendonca_query_2020}
Mendon{\c c}a, V., Sardinha, A., Coheur, L., Santos, A.L.: Query {{Strategies}}, {{Assemble}}! {{Active Learning}} with {{Expert Advice}} for {{Low-resource Natural Language Processing}}. In: 2020 {{IEEE International Conference}} on {{Fuzzy Systems}} ({{FUZZ-IEEE}}). pp.~1--8 (Jul 2020). \doi{10.1109/FUZZ48607.2020.9177707}

\bibitem{miller_name_2004}
Miller, S., Guinness, J., Zamanian, A.: Name tagging with word clusters and discriminative training. In: {{HLT-NAACL}} 2004 - {{Human Language Technology Conference}} of the {{North American Chapter}} of the {{Association}} for {{Computational Linguistics}}, {{Proceedings}} of the {{Main Conference}}. pp. 337--342 (2004)

\bibitem{mironczuk_recent_2018}
Miro{\'n}czuk, M.M., Protasiewicz, J.: A recent overview of the state-of-the-art elements of text classification. Expert Systems with Applications  \textbf{106},  36--54 (2018)

\bibitem{mo_learning_2017}
Mo, Y., Scott, S., Downey, D.: Learning hierarchically decomposable concepts with active over-labeling. In: Proceedings - {{IEEE International Conference}} on {{Data Mining}}, {{ICDM}}. pp. 340--349 (2017). \doi{10.1109/ICDM.2016.165}

\bibitem{moniz_efficiently_2022}
Moniz, J., Patra, B., Gormley, M.: On {{Efficiently Acquiring Annotations}} for {{Multilingual Models}}. In: Proceedings of the {{Annual Meeting}} of the {{Association}} for {{Computational Linguistics}}. vol.~2, pp. 69--85 (2022)

\bibitem{munnSystematicReviewScoping2018}
Munn, Z., Peters, M.D.J., Stern, C., Tufanaru, C., McArthur, A., Aromataris, E.: Systematic review or scoping review? {{Guidance}} for authors when choosing between a systematic or scoping review approach. BMC Medical Research Methodology  \textbf{18}(1), ~143 (Nov 2018). \doi{10.1186/s12874-018-0611-x}

\bibitem{neto_deep_2021}
Neto, J., Faleiros, T.: Deep {{Active-Self Learning Applied}} to {{Named Entity Recognition}}. In: Lecture {{Notes}} in {{Computer Science}} (Including Subseries {{Lecture Notes}} in {{Artificial Intelligence}} and {{Lecture Notes}} in {{Bioinformatics}}). vol. 13074 LNAI, pp. 405--418 (2021). \doi{10.1007/978-3-030-91699-2\_28}

\bibitem{nguyen_active_2013}
Nguyen, V., Lee, W., Ye, N., Chai, K., Chieu, H.: Active learning for probabilistic hypotheses using the maximum {{Gibbs}} error criterion. In: Advances in {{Neural Information Processing Systems}} (2013)

\bibitem{ni_fast_2015}
Ni, J., Delaney, B., Florian, R.: Fast {{Model Adaptation}} for {{Automated Section Classification}} in {{Electronic Medical Records}}. In: Studies in {{Health Technology}} and {{Informatics}}. vol.~216, pp. 35--39 (2015). \doi{10.3233/978-1-61499-564-7-35}

\bibitem{olsson_privacy_2009}
Olsson, F.: On privacy preservation in text and document-based active learning for named entity recognition. In: International {{Conference}} on {{Information}} and {{Knowledge Management}}, {{Proceedings}}. pp. 53--60 (2009). \doi{10.1145/1651449.1651460}

\bibitem{olsson_intrinsic_2009}
Olsson, F., Tomanek, K.: An intrinsic stopping criterion for committee-based active learning. In: {{CoNLL}} 2009 - {{Proceedings}} of the {{Thirteenth Conference}} on {{Computational Natural Language Learning}}. pp. 138--146 (2009). \doi{10.3115/1596374.1596398}

\bibitem{pradhan_knowledge_2020}
Pradhan, A., Todi, K., Selvarasu, A., Sanyal, A.: Knowledge {{Graph Generation}} with {{Deep Active Learning}}. In: Proceedings of the {{International Joint Conference}} on {{Neural Networks}} (2020). \doi{10.1109/IJCNN48605.2020.9207515}

\bibitem{radmard_subsequence_2021}
Radmard, P., Fathullah, Y., Lipani, A.: Subsequence {{Based Deep Active Learning}} for {{Named Entity Recognition}}. In: Proceedings of the 59th {{Annual Meeting}} of the {{Association}} for {{Computational Linguistics}} and the 11th {{International Joint Conference}} on {{Natural Language Processing}} ({{Volume}} 1: {{Long Papers}}). pp. 4310--4321. {Association for Computational Linguistics}, {Online} (Aug 2021). \doi{10.18653/v1/2021.acl-long.332}

\bibitem{ren_survey_2022}
Ren, P., Xiao, Y., Chang, X., Huang, P.Y., Li, Z., Gupta, B.B., Chen, X., Wang, X.: A {{Survey}} of {{Deep Active Learning}}. ACM Computing Surveys  \textbf{54}(9),  1--40 (Dec 2022). \doi{10.1145/3472291}

\bibitem{saha_active_2012}
Saha, S., Ekbal, A., Verma, M., Sikdar, U., Poesio, M.: Active learning technique for biomedical named entity extraction. In: {{ACM International Conference Proceeding Series}}. pp. 835--841 (2012). \doi{10.1145/2345396.2345532}

\bibitem{sapci_focusing_2023}
{\c S}apci, A., Kemik, H., Yeniterzi, R., Tastan, O.: Focusing on potential named entities during active label acquisition. Natural Language Engineering  (2023). \doi{10.1017/S1351324923000165}

\bibitem{settles_active_2009}
Settles, B.: Active {{Learning Literature Survey}} p.~67 (2009)

\bibitem{settles_analysis_2008}
Settles, B., Craven, M.: An analysis of active learning strategies for sequence labeling tasks. In: Proceedings of the 2008 Conference on Empirical Methods in Natural Language Processing. pp. 1070--1079 (2008)

\bibitem{shardlow_text_2019}
Shardlow, M., Ju, M., Li, M., O'Reilly, C., Iavarone, E., McNaught, J., Ananiadou, S.: A {{Text Mining Pipeline Using Active}} and {{Deep Learning Aimed}} at {{Curating Information}} in {{Computational Neuroscience}}. Neuroinformatics  \textbf{17}(3),  391--406 (2019). \doi{10.1007/s12021-018-9404-y}

\bibitem{sharma_named_2022}
Sharma, A., {Amrita}, Chakraborty, S., Kumar, S.: Named {{Entity Recognition}} in {{Natural Language Processing}}: {{A Systematic Review}}. In: Gupta, D., Khanna, A., Kansal, V., Fortino, G., Hassanien, A.E. (eds.) Proceedings of {{Second Doctoral Symposium}} on {{Computational Intelligence}}. pp. 817--828. Advances in {{Intelligent Systems}} and {{Computing}}, {Springer}, {Singapore} (2022). \doi{10.1007/978-981-16-3346-1\_66}

\bibitem{shelmanov_active_2021}
Shelmanov, A., Puzyrev, D., Kupriyanova, L., Belyakov, D., Larionov, D., Khromov, N., Kozlova, O., Artemova, E., Dylov, D.V., Panchenko, A.: Active {{Learning}} for {{Sequence Tagging}} with {{Deep Pre-trained Models}} and {{Bayesian Uncertainty Estimates}} (Feb 2021)

\bibitem{shen_deep_2017}
Shen, Y., Yun, H., Lipton, Z., Kronrod, Y., Anandkumar, A.: Deep active learning for named entity recognition. In: Proceedings of the 2nd {{Workshop}} on {{Representation Learning}} for {{NLP}}, {{Rep4NLP}} 2017 at the 55th {{Annual Meeting}} of the {{Association}} for {{Computational Linguistics}}, {{ACL}} 2017. pp. 252--256 (2017)

\bibitem{shen_deep_2018}
Shen, Y., Yun, H., Lipton, Z.C., Kronrod, Y., Anandkumar, A.: Deep {{Active Learning}} for {{Named Entity Recognition}} (Feb 2018)

\bibitem{shrivastava_iseql_2020}
Shrivastava, A., Heer, J.: {{ISeqL}}. In: International {{Conference}} on {{Intelligent User Interfaces}}, {{Proceedings IUI}}. pp. 43--54 (2020). \doi{10.1145/3377325.3377503}

\bibitem{siddhant_deep_2018}
Siddhant, A., Lipton, Z.C.: Deep {{Bayesian Active Learning}} for {{Natural Language Processing}}: {{Results}} of a {{Large-Scale Empirical Study}} (Sep 2018)

\bibitem{simpson_bayesian_2019}
Simpson, E., Gurevych, I.: A {{Bayesian}} approach for sequence tagging with crowds. In: {{EMNLP-IJCNLP}} 2019 - 2019 {{Conference}} on {{Empirical Methods}} in {{Natural Language Processing}} and 9th {{International Joint Conference}} on {{Natural Language Processing}}, {{Proceedings}} of the {{Conference}}. pp. 1093--1104 (2019)

\bibitem{sintayehu_named_2021}
Sintayehu, H., Lehal, G.S.: Named entity recognition: A semi-supervised learning approach. International Journal of Information Technology  \textbf{13}(4),  1659--1665 (Aug 2021). \doi{10.1007/s41870-020-00470-4}

\bibitem{skeppstedt_visualising_2020}
Skeppstedt, M., Rzepka, R., Araki, K., Kerren, A.: Visualising and evaluating the effects of combining active learning with word embedding features. In: Proceedings of the 15th {{Conference}} on {{Natural Language Processing}}, {{KONVENS}} 2019. pp. 91--100 (2020)

\bibitem{son_jointly_2022}
Son, N.H., Yu, H.M., Nguyen, T.A.D., Nguyen, M.T.: Jointly {{Learning Span Extraction}} and {{Sequence Labeling}} for {{Information Extraction}} from {{Business Documents}}. In: 2022 {{International Joint Conference}} on {{Neural Networks}} ({{IJCNN}}). pp.~1--8 (Jul 2022). \doi{10.1109/IJCNN55064.2022.9892779}

\bibitem{tang_towards_2022}
Tang, S., Liu, H., Almatared, M., Abudayyeh, O., Lei, Z., Fong, A.: Towards {{Automated Construction Quantity Take-Off}}: {{An Integrated Approach}} to {{Information Extraction}} from {{Work Descriptions}}. Buildings  \textbf{12}(3) (2022). \doi{10.3390/buildings12030354}

\bibitem{tang_learning_2021}
Tang, X., Wu, S., Chen, G., Chen, K., Shou, L.: Learning to {{Label}} with {{Active Learning}} and {{Reinforcement Learning}}. In: Lecture {{Notes}} in {{Computer Science}} (Including Subseries {{Lecture Notes}} in {{Artificial Intelligence}} and {{Lecture Notes}} in {{Bioinformatics}}). vol. 12682 LNCS, pp. 549--557 (2021). \doi{10.1007/978-3-030-73197-7\_36}

\bibitem{tchoua_active_2019}
Tchoua, R., Ajith, A., Hong, Z., Ward, L., Chard, K., Audus, D., Patel, S., De~Pablo, J., Foster, I.: Active learning yields better training data for scientific named entity recognition. In: Proceedings - {{IEEE}} 15th {{International Conference}} on {{eScience}}, {{eScience}} 2019. pp. 126--135 (2019). \doi{10.1109/eScience.2019.00021}

\bibitem{tjong_kim_sang_introduction_2003}
Tjong Kim~Sang, E.F., De~Meulder, F.: Introduction to the {{CoNLL-2003 Shared Task}}: {{Language-Independent Named Entity Recognition}}. In: Proceedings of the {{Seventh Conference}} on {{Natural Language Learning}} at {{HLT-NAACL}} 2003. pp. 142--147 (2003)

\bibitem{tomanek_approximating_2008}
Tomanek, K., Hahn, U.: Approximating learning curves for active-learning-driven annotation. In: Proceedings of the 6th {{International Conference}} on {{Language Resources}} and {{Evaluation}}, {{LREC}} 2008. pp. 1319--1324 (2008)

\bibitem{tomanek_reducing_2009}
Tomanek, K., Hahn, U.: Reducing class imbalance during active learning for named entity annotation. In: K-{{CAP}}'09 - {{Proceedings}} of the 5th {{International Conference}} on {{Knowledge Capture}}. pp. 105--112 (2009). \doi{10.1145/1597735.1597754}

\bibitem{tomanek_annotation_2010}
Tomanek, K., Hahn, U.: Annotation time stamps - {{Temporal}} metadata from the linguistic annotation process. In: Proceedings of the 7th {{International Conference}} on {{Language Resources}} and {{Evaluation}}, {{LREC}} 2010. pp. 2516--2521 (2010)

\bibitem{tomanek_proper_2009}
Tomanek, K., Laws, F., Hahn, U., Sch{\"u}tze, H.: On proper unit selection in active learning: Co-selection effects for named entity recognition. In: Proceedings of the {{NAACL HLT}} 2009 Workshop on Active Learning for Natural Language Processing. pp. 9--17. {{HLT}} '09, {Association for Computational Linguistics}, {USA} (2009)

\bibitem{tran_hybrid_2017}
Tran, V., Hoang, D., Nguyen, N., Hwang, D.: A hybrid method for named entity recognition on tweet streams. In: Lecture {{Notes}} in {{Computer Science}} (Including Subseries {{Lecture Notes}} in {{Artificial Intelligence}} and {{Lecture Notes}} in {{Bioinformatics}}). vol. 10191 LNAI, pp. 258--268 (2017). \doi{10.1007/978-3-319-54472-4\_25}

\bibitem{tran_combination_2017}
Tran, V., Nguyen, N., Fujita, H., Hoang, D., Hwang, D.: A combination of active learning and self-learning for named entity recognition on {{Twitter}} using conditional random fields. Knowledge-Based Systems  \textbf{132},  179--187 (2017). \doi{10.1016/j.knosys.2017.06.023}

\bibitem{triccoPRISMAExtensionScoping2018}
Tricco, A.C., Lillie, E., Zarin, W., O'Brien, K.K., Colquhoun, H., Levac, D., Moher, D., Peters, M.D., Horsley, T., Weeks, L., Hempel, S., Akl, E.A., Chang, C., McGowan, J., Stewart, L., Hartling, L., Aldcroft, A., Wilson, M.G., Garritty, C., Lewin, S., Godfrey, C.M., Macdonald, M.T., Langlois, E.V., {Soares-Weiser}, K., Moriarty, J., Clifford, T., Tun{\c c}alp, {\"O}., Straus, S.E.: {{PRISMA Extension}} for {{Scoping Reviews}} ({{PRISMA-ScR}}): {{Checklist}} and {{Explanation}}. Annals of Internal Medicine  \textbf{169}(7),  467--473 (Oct 2018). \doi{10.7326/M18-0850}

\bibitem{uzuner_2010_2011}
Uzuner, {\"O}., South, B.R., Shen, S., DuVall, S.L.: 2010 i2b2/{{VA}} challenge on concepts, assertions, and relations in clinical text. Journal of the American Medical Informatics Association : JAMIA  \textbf{18}(5),  552--556 (2011). \doi{10.1136/amiajnl-2011-000203}

\bibitem{van_nguyen_famie_2022}
Van~Nguyen, M., Ngo, N., Min, B., Nguyen, T.: {{FAMIE}}: {{A Fast Active Learning Framework}} for {{Multilingual Information Extraction}}. In: {{NAACL}} 2022 - 2022 {{Conference}} of the {{North American Chapter}} of the {{Association}} for {{Computational Linguistics}}: {{Human Language Technologies}}, {{Proceedings}} of the {{Demonstrations Session}}. pp. 131--139 (2022)

\bibitem{vaswaniAttentionAllYou2023}
Vaswani, A., Shazeer, N., Parmar, N., Uszkoreit, J., Jones, L., Gomez, A.N., Kaiser, L., Polosukhin, I.: Attention {{Is All You Need}} (Aug 2023). \doi{10.48550/arXiv.1706.03762}

\bibitem{vaswani_attention_2023}
Vaswani, A., Shazeer, N., Parmar, N., Uszkoreit, J., Jones, L., Gomez, A.N., Kaiser, L., Polosukhin, I.: Attention {{Is All You Need}} (Aug 2023). \doi{10.48550/arXiv.1706.03762}

\bibitem{veerasekharreddy_named_2022}
Veerasekharreddy, B., Rao, K., Koppula, N.: Named {{Entity Recognition}} using {{CRF}} with {{Active Learning Algorithm}} in {{English Texts}}. In: 6th {{International Conference}} on {{Electronics}}, {{Communication}} and {{Aerospace Technology}}, {{ICECA}} 2022 - {{Proceedings}}. pp. 1041--1044 (2022). \doi{10.1109/ICECA55336.2022.10009592}

\bibitem{verma_ensemble_2013}
Verma, M., Sikdar, U., Saha, S., Ekbal, A.: Ensemble based active annotation for biomedical named entity recognition. In: Proceedings of the 2013 {{International Conference}} on {{Advances}} in {{Computing}}, {{Communications}} and {{Informatics}}, {{ICACCI}} 2013. pp. 973--978 (2013). \doi{10.1109/ICACCI.2013.6637308}

\bibitem{wei_cost-aware_2019}
Wei, Q., Chen, Y., Salimi, M., Denny, J., Mei, Q., Lasko, T., Chen, Q., Wu, S., Franklin, A., Cohen, T., Xu, H.: Cost-aware active learning for named entity recognition in clinical text. Journal of the American Medical Informatics Association  \textbf{26}(11),  1314--1322 (2019). \doi{10.1093/jamia/ocz102}

\bibitem{yao_looking_2020}
Yao, J., Dou, Z., Nie, J., Wen, J.: Looking {{Back}} on the {{Past}}: {{Active Learning}} with {{Historical Evaluation Results}}. IEEE Transactions on Knowledge and Data Engineering  (2020). \doi{10.1109/TKDE.2020.3045816}

\bibitem{zaratiana_gnner_2022}
Zaratiana, U., Tomeh, N., Holat, P., Charnois, T.: {{GNNer}}: {{Reducing Overlapping}} in {{Span-based NER Using Graph Neural Networks}}. In: Proceedings of the 60th {{Annual Meeting}} of the {{Association}} for {{Computational Linguistics}}: {{Student Research Workshop}}. pp. 97--103. {Association for Computational Linguistics}, {Dublin, Ireland} (2022). \doi{10.18653/v1/2022.acl-srw.9}

\bibitem{zhan_comparative_2022}
Zhan, X., Wang, Q., Huang, K.h., Xiong, H., Dou, D., Chan, A.B.: A {{Comparative Survey}} of {{Deep Active Learning}} (Jul 2022)

\bibitem{zhang_survey_2022}
Zhang, Z., Strubell, E., Hovy, E.: A survey of active learning for natural language processing. arXiv preprint arXiv:2210.10109  (2022)

\bibitem{zheng_opentag_2018}
Zheng, G., Mukherjee, S., Dong, X., Li, F.: {{OpenTag}}: {{Open}} aribute value extraction from product profiles. In: Proceedings of the {{ACM SIGKDD International Conference}} on {{Knowledge Discovery}} and {{Data Mining}}. pp. 1049--1058 (2018). \doi{10.1145/3219819.3219839}

\bibitem{zhong_chinese_2014}
Zhong, Z., Liu, F., Wu, Y., Wu, J.: Chinese named entity recognition combined active learning with self-training. Guofang Keji Daxue Xuebao/Journal of National University of Defense Technology  \textbf{36}(4),  82--88 (2014). \doi{10.11887/j.cn.201404015}

\bibitem{zhou_mtaal_2021}
Zhou, B., Cai, X., Zhang, Y., Guo, W., Yuan, X.: {{MTAAL}}: {{Multi-Task Adversarial Active Learning}} for {{Medical Named Entity Recognition}} and {{Normalization}}. In: 35th {{AAAI Conference}} on {{Artificial Intelligence}}, {{AAAI}} 2021. vol.~16, pp. 14586--14593 (2021)

\bibitem{zhou_progress_2020}
Zhou, M., Duan, N., Liu, S., Shum, H.Y.: Progress in neural {{NLP}}: Modeling, learning, and reasoning. Engineering  \textbf{6}(3),  275--290 (2020)

\bibitem{zhou_active_2023}
Zhou, S., Liang, S., Yang, Q., Jiang, W., He, Y., Li, Y.: Active {{Learning Based Labeling Method}} for~{{Fault Disposal Pre-plans}}. In: Advances and {{Trends}} in {{Artificial Intelligence}}. {{Theory}} and {{Applications}}. pp. 377--382 (2023)

\bibitem{zhuo_red_2023}
Zhuo, T.Y., Huang, Y., Chen, C., Xing, Z.: Red teaming {{ChatGPT}} via {{Jailbreaking}}: {{Bias}}, {{Robustness}}, {{Reliability}} and {{Toxicity}} (May 2023). \doi{10.48550/arXiv.2301.12867}

\end{thebibliography}
\end{document}